\documentclass[a4paper,conference]{IEEEtran}
\IEEEoverridecommandlockouts
\usepackage{cite}
\usepackage{amsmath,amssymb,amsfonts}
\usepackage{graphicx}
\usepackage{textcomp}
\usepackage{xcolor}

\usepackage{subfigure} %test
\usepackage{algorithm}
\usepackage{algpseudocode}
\usepackage{graphics}
\usepackage{epsfig}
\usepackage{url}      %Required
\usepackage{booktabs}
\usepackage{multirow}

\def\BibTeX{{\rm B\kern-.05em{\sc i\kern-.025em b}\kern-.08em
    T\kern-.1667em\lower.7ex\hbox{E}\kern-.125emX}}
\begin{document}

\title{Semantics-Guided Representation Learning\\with Applications to Visual Synthesis\\}

\author{\IEEEauthorblockN{Jia-Wei Yan$^{1}$,
Ci-Siang Lin$^{1,2}$,
Fu-En Yang$^{1,2}$,
Yu-Jhe Li$^{3}$,and
Yu-Chiang Frank Wang$^{1,2}$}
\IEEEauthorblockA{$^{1}$Graduate Institute of Communication Engineering, National Taiwan University, Taiwan \\
$^{2}$ASUS Intelligent Cloud Services, Taiwan \\
$^{3}$Robotics Institute, Carnegie Mellon University,
Pittsburgh, USA \\ 
Email: \{r06942033, d08942011, r07942077, ycwang\}@ntu.edu.tw, yujheli@cs.cmu.edu}
}

\maketitle

\begin{abstract}

Learning interpretable and interpolatable latent representations has been an emerging research direction, allowing researchers to understand and utilize the derived latent space for further applications such as visual synthesis or recognition. While most existing approaches derive an interpolatable latent space and induces smooth transition in image appearance, it is still not clear how to observe desirable representations which would contain semantic information of interest. In this paper, we aim to learn meaningful representations and simultaneously perform semantic-oriented and visually-smooth interpolation. To this end, we propose an angular triplet-neighbor loss (ATNL) that enables learning a latent representation whose distribution matches the semantic information of interest.
With the latent space guided by ATNL, we further utilize spherical semantic interpolation for generating semantic warping of images, allowing synthesis of desirable visual data. Experiments on MNIST and CMU Multi-PIE datasets qualitatively and quantitatively verify the effectiveness of our method.

\end{abstract}


\section{Introduction}

% Representation learning
In the field of machine learning and computer vision, learning interpretable representations has been active research topic~\cite{tripletloss,tripletcenterloss,infogan_nips2016,betavae_iclr2017}, benefiting a range of applications such as visual classification, retrieval and synthesis. Most existing approaches approach such tasks by deriving \textit{interpolatable} latent spaces, resulting in smooth transition in visual appearances. However, it is still not clear how to observe desirable  representations which would contain semantic information of interest. In this paper, we address a novel task of learning semantically interpretable yet interpolatable representations with applications to visual analysis.

\begin{figure}[t]
  \centering
  \includegraphics[width=\linewidth]{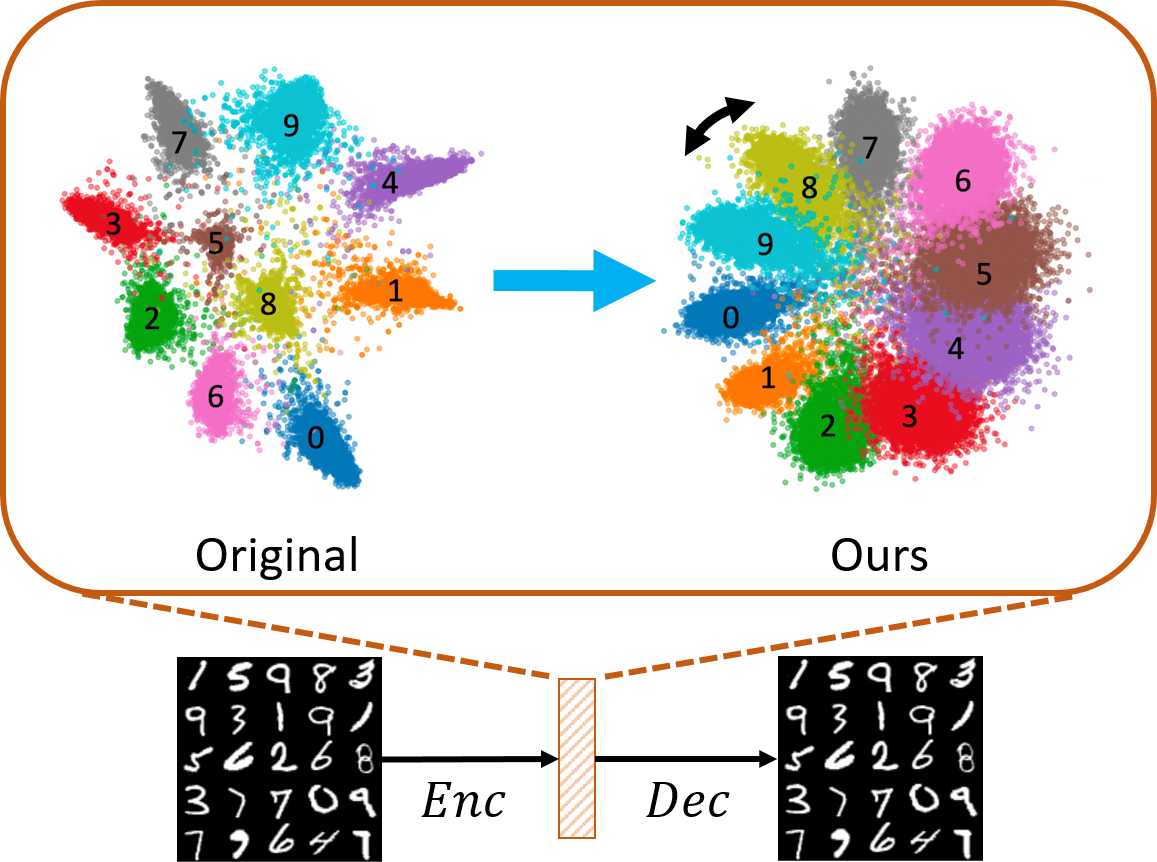}
  \caption{To manipulate the latent representations which semantically match the images in numerical order, we aim at learning latent representations which properly describe the associated semantic information. Take MNIST as examples, while VAE learns latent representation distributions which correlate with visual appearances of images (left), ours (right) follows the numerical order.
  }
  \label{fig:idea}
\end{figure}

Approaches for representation learning can be divided into supervised and unsupervised settings.
% supervised
For supervised learning, take face recognition for example, one needs to extract discriminative representations from input images of different identities. Previous works like~\cite{tripletloss,centerloss,sphereface_cpvr2017,cosfae_cvpr2018,arcface,ringloss} use various constraints or regularizers for making features of the same person to be simialar, while those of different people to be distant from each other. As for retrieval tasks like~\cite{tripletcenterloss,atcl_aaai2019}, one needs to derive effective representations, so that intra and inter-category properties can be observed. 

% unsupervised

On the other hand, to address unsupervised representation learning, a number of generative models have been proposed. For example, autoencoder (AE)-based models like~\cite{denoisingae_2010,vae_2013} are presented to ensure the derived latent features to sufficiently represent data  information. Goodfellow~\textit{et al.}~\cite{gan_nips2014} propose Generative Adversarial Networks (GANs), which further exhibits promising ability in image generation. To extend GANs for representation disentanglement, recent works like~\cite{infogan_nips2016,unit_nips2017,munit_eccv2018,drit_eccv2018} decompose the latent representation into distinct parts, with the goal of distilling the attributes of interest (e.g., pose, expression, etc.).

%transition
However, it is worth noting that existing representation learning works generally focus on learning representation for either classification or synthesis guarantees. It is not clear how to observe desirable representations which would contain proper semantic information. Berthelot \textit{et al.}~\cite{acai_iclr2019} leverages regularization strategy with the specific goal of encouraging improved interpolation ability, showing solid ability in performing interpolation for producing meaningful representation. However, to the best of our knowledge, most existing works are not able to generate latent representations which semantically match the visual data  of interest via interpolation. Take MNIST~\cite{mnist} in Figure~\ref{fig:idea} for example, when interpolating autoencoder-based latent features of images between digits 0 and 2, instead of semantic warping of images following the numerical order (i.e., digits $0 \rightarrow 1 \rightarrow 2$), the output images would visually and gradually warp from digits 0 to 6, and then finally to 2.

% our proposed
To address the above novel yet challenging problem, we propose a novel Angular Triplet-Neighbor Loss (ATNL) for deriving latent representations whose distribution would match the semantic information. As depicted in Figure~\ref{fig:idea}, we aim to learn a latent representation for visual data with distribution semantically fitting the associated learning task. Inspired by FaceNet~\cite{tripletloss} and SphereFace~\cite{sphereface_cpvr2017}, we propose unique semantics-guided loss with novel spherical semantic interpolation techniques, which allows our model to generate semantically desirable image outputs. In addition to standard quantitative and qualitative evaluation on our derive latent representations, we uniquely apply our model as a data hallucination technique, which synthesizes visual data of particular categeries and address few-shot learning tasks.

%Contribution
We now highlight the contributions of our work as follows:
\begin{itemize}
    \item We are among the first to explore desirable semantic distribution of latent representations, resulting in interpretable and interpolatable representations.
    \item We propose an Angular Triplet-Neighbor Loss (ATNL), followed by spherical semantic interpolation, which utilizes task-oriented semantic information for representation learning.
    \item In addition to qualitative and quantitative evaluation on our learned representation, we further extend our learning strategy as a data hallucination technique, which is successfully applied for few-shot image classification.
\end{itemize}

\section{Related Work}

\subsection{Unsupervised Representation Learning}
When ground truth labels are not observed, one needs to perform unsupervised learning to derive deep feature representations. With data recovery guarantees, autoencoder (AE) and variational autorencoder (VAE)~\cite{vae_2013} have been proposed. While the generative model of Generative Adversarial Networks (GANs)~\cite{gan_nips2014} is able to produce realistic visual data, it lacks the ability to extract particular features corresponding to the attributes of interest. To allow disentanglement of such features, info-GAN~\cite{infogan_nips2016} maximizes mutual information between latent variables and data variation. Bidirectional GAN (BiGAN)~\cite{bigan} achieves representation learning via incorporating an encoder to a vanilla GAN. Based on the architectures of VAE and GAN, recent models of DRIT~\cite{drit_eccv2018} and MUNIT~\cite{munit_eccv2018} extend UNIT~\cite{unit_nips2017} for translating images across domains, resulting domain-invariant features as the derived latent representation. Berthelot~\textit{et al.}~\cite{acai_iclr2019} further extend AE with adversarial regularizer, and propose Adversarially Constrained Autoencoder Interpolation (ACAI) which encourage the images recovered by latent representation interpolation to be realistic.

\subsection{Supervised Representation Learning}

To realized supervised representation learning, contrastive loss~\cite{contrastiveloss} and triplet loss~\cite{tripletloss} have been applied to increase Euclidean distance between different category for discriminative feature embedding. Wenl~\textit{et al.}~\cite{centerloss} propose the center loss to better cluster deep features of each identity, aiming at reducing intra-class variance. Liu~\textit{et al.}~\cite{lsoftmax} propose a large margin softmax (L-Softmax) by adding angular constraints to each identity to improve feature discrimination. Angular softmax (A-Softmax)~\cite{sphereface_cpvr2017} further improves L-Softmax by normalizing the weights of last fully connected layer for improved face recognition. Other loss functions~\cite{tripletcenterloss,ringloss,cosfae_cvpr2018,arcface,atcl_aaai2019} extended from the above losses have demonstrated promising performances on the associated learning tasks. Nevertheless, even under supervised learning settings, existing works on representation learning cannot easily generate or manipulate latent representations which semantically match the images of interest via interpolation (see~\ref{fig:idea} for example). %This is the reason 
\section{Semantics-Guided Representation Learning}

% \begin{figure}
%   \centering
%   \includegraphics[width=\linewidth]{img/architecture.png}
%   \caption{Overview of our framework. The introduced Angular Triplet-Neighbor Loss (ATNL) is applied to the latent space in the naive VAE, which consists of the encoder $E$ and the decoder $D$. Note that $D(z)$ denote the reconstructed images with latent feature $z$ as input, while the distribution of $z$ is approximated to $\mathcal{N}(0,1)$.}
%   \label{fig:archi}
% \end{figure}

\begin{figure}[t]
\vspace{2mm}
  \centering
  \includegraphics[width=\linewidth]{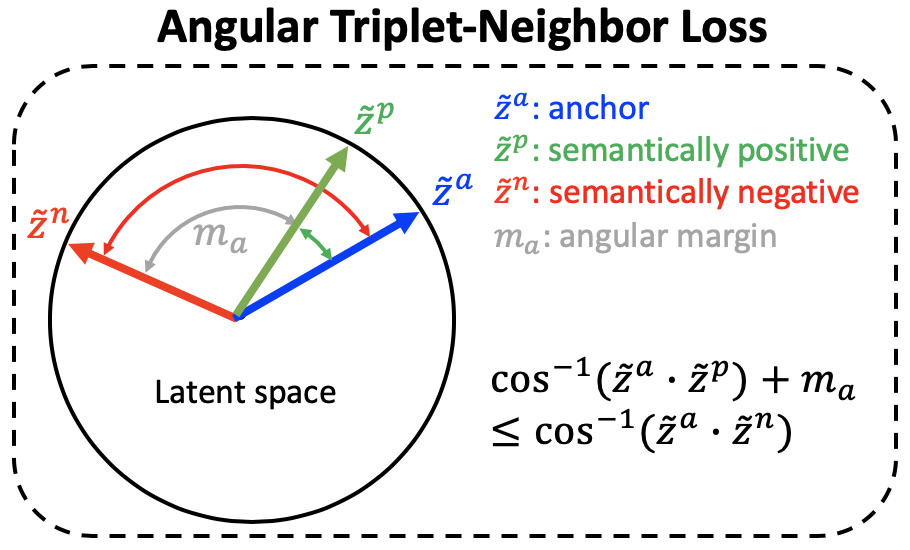}
  \caption{Our Angular Triplet-Neighbor Loss (ATNL) is developed to enforce the positive normalized feature $\tilde{z}^p$ become close to the anchor feature $\tilde{z}^a$, while the negative feature $\tilde{z}^n$ would kept away from the positive one by an angular margin $m_a$.}
  \label{fig:atnl}
\end{figure}

\subsection{VAE for Representation Learning}

The use of variational autoencoders (VAE) for image recovery (i.e., output $D(z)$) while resulting in latent features for describing the input images. Its an encoder $E$ to map an input image $x$ into a latent subspace with latent code $z$, whose distribution fits the Gaussian distribution $\mathcal{N}(0,1)$. On the other hand, the decoder $D$ in VAE is designed to reconstruct the input image from $z$. The L1 loss is typically applied on the output image $D(z)$ as the reconstruction loss of VAE, which measures the difference between $x$ and $D(z)$:

\begin{equation}
\label{eq_1}
L_{rec} = |D(z)-x|.
\end{equation}

\noindent And, the KL divergence is calculated to fit the observed latent representation distribution to $\mathcal{N}(0,1)$, i.e.,

\begin{equation}
\label{eq_2}
L_{KL} = \mathbb{E}[KL(P(z)||\mathcal{N}(0,1))].
\end{equation}

Since VAE focuses on deriving continuous latent subspaces for representing image features with recovery guarantees, it does not necessarily describe such features based on their semantic information. As illustrated in Figure~\ref{fig:atnl} and detailed later, we introduce an novel Angular Triplet-Neighbor Loss (ATNL) at the latent representation of VAE. This not only fits data latent distribution as the aforementioned Gaussian one, it additionally allows the resulting latent features to exhibit their semantic information of interest.

%------------------------------------------

\subsection{Angular Triplet-Neighbor Loss (ATNL)}
\label{subsec:ATNL}

To overcome the aforementioned limitations in current VAE-based deep learning models for image generation and understanding, we introduce a novel Angular Triplet-Neighbor Loss (ATNL) to existing VAE frameworks. 
Inspired by FaceNet~\cite{tripletloss}, we extend the original triplet loss to better describe and match task-specific semantic information in derived latent space. To start, we have the $i$th input image as the anchor (with latent code $z_{i}^{a}$). Instead of simply viewing image data of the same and different categories as positive and negative samples as FaceNet does, we choose to $z_{i}^{p}$ as the \textit{semantically-positive} samples, whose categories are the same or semantically similar to that of $z_{i}^{a}$. As for the samples of the remaining categories, they will be simply viewed as the negative ones $z_{i}^{n}$. The above semantic relationship between samples would be determined based on the task of interest. Take handwritten digit classification for example, given an anchor sample $z^{a}$ of digit 5, its semantically-positive samples $z^{p}$ would be those of digits 4 and 6. And, the samples in the remaining categories (i.e., digits other than 4, 5, and 6) would be semantically-negative ones $z^{n}$.

\begin{figure}[t]
\vspace{2mm}
  \begin{center}
  \includegraphics[width=\linewidth]{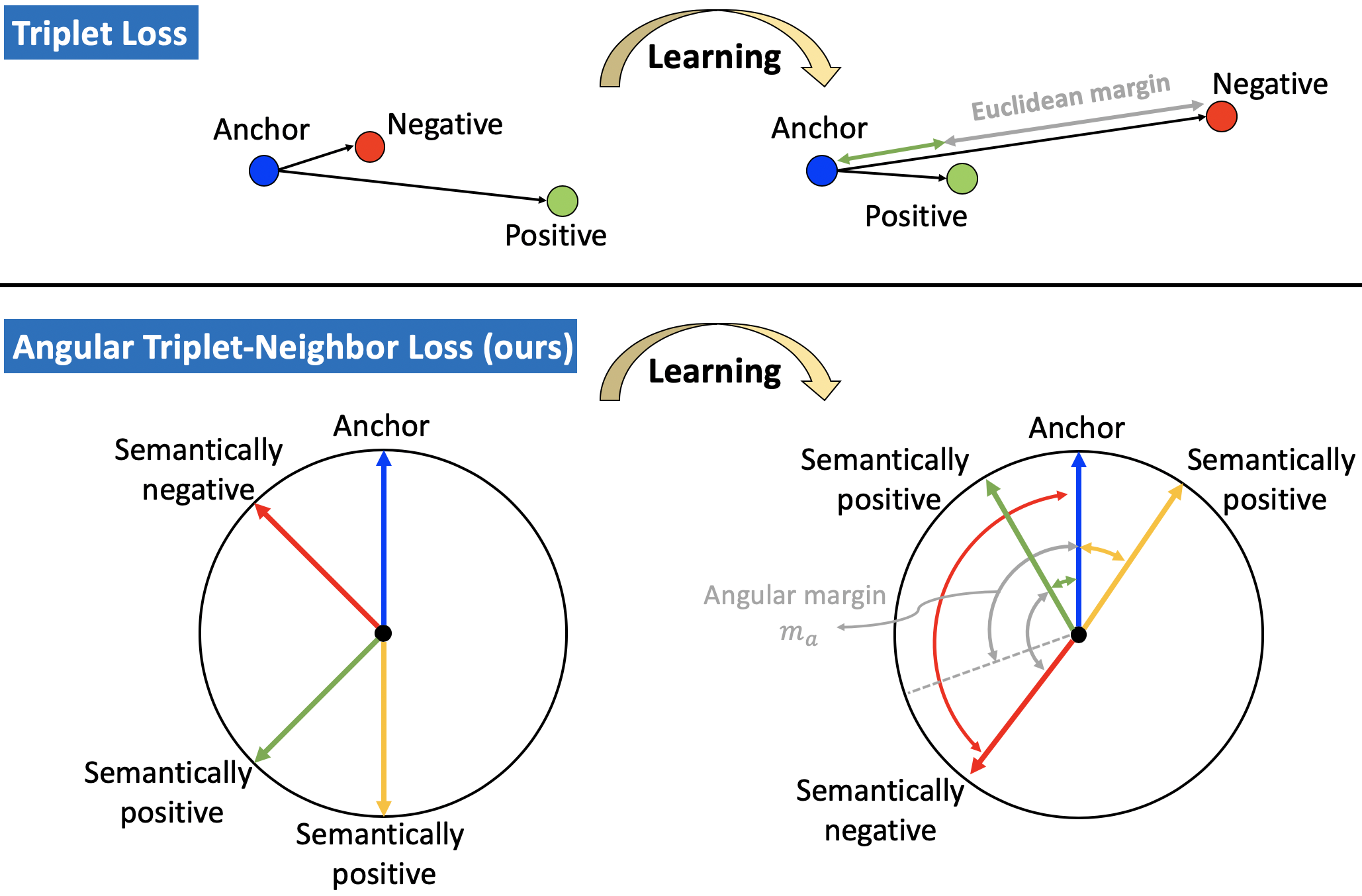}
  \caption{Comparisons between the standard triplet loss and our ATNL. The triplet loss minimizes the Euclidean distance between the anchor and the positive instance, while maximizing the distance between the anchor and a negative one. The difference of the two distances are enforced to kept a pre-defined margin. Our ATNL utilizes angular distances and semantically-positive/negative samples, which preserve data discriminativity with a normalized margin.
%   and redefine semantically-positive/negative which is not only singular, it may be plural.
  }
  \label{fig:definition}
  \end{center}
\end{figure}

With semantics-dependent positive and negative samples determined, we advance the angular-based metrics to calculate the ATNL. As depicted in Figure~\ref{fig:atnl}, we normalize the latent code/features of input images, map them onto a unit hypersphere (i.e., $||z|| =1$), and calculate the following equation for satisfying ATNL:

\begin{equation}
\label{eq_3}
arccos(\tilde{z}_{i}^{a} \cdot \tilde{z}_{i}^{p}) + m_a \leq arccos(\tilde{z}_{i}^{a} \cdot \tilde{z}_{i}^{n}),
\end{equation}
\begin{equation*}
\forall (\tilde{z}_{i}^{a}, \tilde{z}_{i}^{p}, \tilde{z}_{i}^{n}) \in T_{p} .
\end{equation*}

\noindent where $\tilde{z}$ and ($\cdot$) denote normalized features and the associate inner product operation, respectively. We have $m_a$ as the angular margin, which is a data-dependent hyperparameter to select. 

It is worth noting that, the use of angular-based metrics requires one to normalize the data onto a unit hypersphere. As noted by ATCL~\cite{atcl_aaai2019} and SphereFace~\cite{sphereface_cpvr2017}, this provides geometric interpretation for the projected data. More precisely, the separation (i.e., margin $m_a$) between data would be in a fixed range of [0, $\pi$]. This is how we adopt angular-based triplet loss instead of traditional triplet loss in our work.

Figure~\ref{fig:definition} illustrates the difference between the standard triplet loss and our ATNL. Later we will explain the unit-sphere normalization is necessary in our model, which makes semantics-guided image generation possible.

To learn the encoder $E$ for VAEs, we thus calculate ATNL at the latent space by:
\begin{equation}
\label{eq_4}
L_{ATN} = \sum_{i=1}^{N} max(arccos(\tilde{z}_{i}^{a} \cdot \tilde{z}_{i}^{p}) - arccos(\tilde{z}_{i}^{a} \cdot \tilde{z}_{i}^{n}) + m_a, 0).
\end{equation}

% triplet selection
Similar to the practical use of triplet loss during training, generating all semantically positive/negative triplet combinations would not be computationally feasible. Alternatively, one possible solution is to select hard triplets, which better contribute to the learning of our models. Thus, we follow~\cite{tripletloss} and utilize \textit{BatchSampler} for online triplet selection, which has been shown to improve learning effectiveness and efficiency.

The total loss of our framework is as follows:

\begin{equation}
\label{ep_7}
L_{total} = \lambda_1 \cdot L_{rec} + \lambda_2 \cdot L_{KL} + \lambda_3 \cdot L_{ATN}.
\end{equation}
where $\lambda_1$, $\lambda_2$ and $\lambda_3$ are hyperparameters of $L_{rec}$, $L_{KL}$ and $L_{ATN}$, respectively.

\subsection{Semantics-Guided Image Generation}

\begin{figure}[t]
\vspace{2mm}
  \begin{center}
  \includegraphics[width=\linewidth]{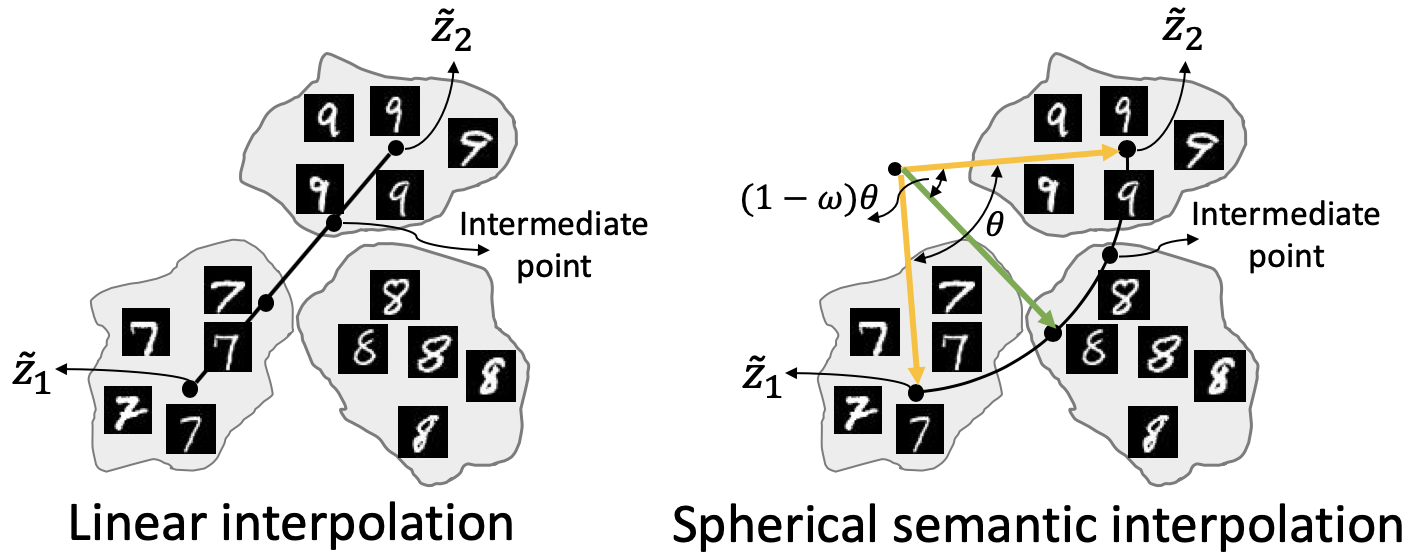}
  \caption{Difference between linear and spherical semantic interpolation on the latent spaces. The former interpolates along a straight line from $\tilde{z}_1$ to $\tilde{z}_2$, while the latter performs along a curve on a unit sphere from $\tilde{z}_1$ to $\tilde{z}_2$. Note that $\theta$ is the angle between $\tilde{z}_1$ and $\tilde{z}_2$ and $\omega$ is a controllable parameter, ranging from 0 to 1.
  }
  \label{fig:interpolation}
  \end{center}
\end{figure}

% sphere semantic interpolation
As described in~\cite{acai_iclr2019}, image generation and manipulation can be realized by interpolating latent features. In general, linear interpolation is the standard technique to apply, which is easily implemented by:

\begin{equation}
\label{ep_5}
\tilde{z}_l(\tilde{z}_1, \tilde{z}_2, \omega) = (1-\omega) \tilde{z}_1 + \omega \tilde{z}_2.
\end{equation}

\noindent where $\tilde{z}_1$ and $\tilde{z}_2$ are the two latent representations of interest, and the weight $\omega$ is a controllable parameter, range from 0 to 1. With $\tilde{z}$, the output image can be produced by $D(\tilde{z}_l)$.

However, image generation via the above linear interpolation simply outputs images with intermediate visual appearance recovered by and between $\tilde{z}_1$ and $\tilde{z}_2$. Take the earlier example of handwritten digit recognition for example, There is no guarantee that images of digit 5 can be interpolated and manipulated by the latent features of digits 4 and 6.

With our ATNL and data normalized onto a unit hypersphere, we perform semantics-guided image generation by advancing \textit{spherical semantic interpolation}, which is performed as follows:

\begin{equation}
\label{ep_6}
\tilde{z}_s(\tilde{z}_1, \tilde{z}_2, \omega) = \frac{sin(1-\omega)\theta}{sin\theta}\tilde{z}_1 + \frac{sin\omega\theta}{sin\theta}\tilde{z}_2.
\end{equation}

\noindent As illustrated shown in Figure~\ref{fig:interpolation}, this operation allows one to generate semantics-guided images from two images of interest, which cannot be easily achieved by standard image recovery via linear interpolation.

% \subsection{Learning of Our Framework}

% Our framework utilizes VAE as the backbone model, with the standard encoder $E$ and decoder $D$ implemented by~\cite{vae_2013}. During the learning of latent representations, the distribution of its associated with the ATNL would have particular semantic information by $E$. Finally, the total loss of our framework as follows:

% \begin{equation}
% \label{ep_7}
% L_{total} = \lambda_1 \cdot L_{rec} + \lambda_2 \cdot L_{KL} + \lambda_3 \cdot L_{ATN}.
% \end{equation}

% \noindent where $\lambda_1$, $\lambda_2$ and $\lambda_3$ are hyper-parameter of $L_{rec}$, $L_{KL}$ and $L_{ATN}$, respectively.

% The pseudo code summarizing the training of our framework is shown in Algorithm 1.
% \input{algorithm.tex}
\section{Experiments}
% We now provide the experimental results in this section. We first present our experimental dataset in Sec.~\ref{exp:data} and the corresponding implementation details in Sec.~\ref{exp:imple}.

\subsection{Datasets}\label{exp:data}
We first consider MNIST~\cite{mnist}, which consists of 60,000/10,000 training/testing handwritten digit images of 10 classes. All images are resized from 28$\times$28 to 32$\times$32 in our experiments. We also use CMU Multi-PIE~\cite{multipie} for our experiments face images. CMU Multi-PIE contains face images with viewpoint, illumination and expression variations. We only use a subset of CMU Multi-PIE with viewpoint variants (24,402 images with 7 viewpoints from -90 degrees to 90 degrees per 30 degrees). All images are resized from 128$\times$128 to 64$\times$64 in our experiments.

\subsection{Implementation Details}\label{exp:imple}
We implemented our model via PyTorch~\cite{pytorch}. The backbone employs VAE~\cite{vae_2013} and the encoder consists of 3 convolution layers with filter size 4$\times$4 followed by three fully connected layers. The decoder consists of 3 fully connected layers followed by 3 convolution layers. For hyper-parameters, we set $\lambda_1 = 10$, $\lambda_2 = 1e-4$, $\lambda_3 = 1$. For MNIST, we set $m_a = 1.2(\approx72^{\circ})$, and we treat the contiguous numbers as semantically positive neighbors while others as the negatives. We also treat digit 0 and digit 9 as neighbors. For Multi-PIE, $m_a = 0.9(\approx50^{\circ})$. We treat contiguous pose of degrees as semantically positive neighbors. It takes 3 hours to train our model for 30 epochs with batch size 256 on a single Nvidia GeForce GTX 1080Ti.

\begin{figure}[t]
\vspace{2mm}
  \centering
  \includegraphics[width=\linewidth]{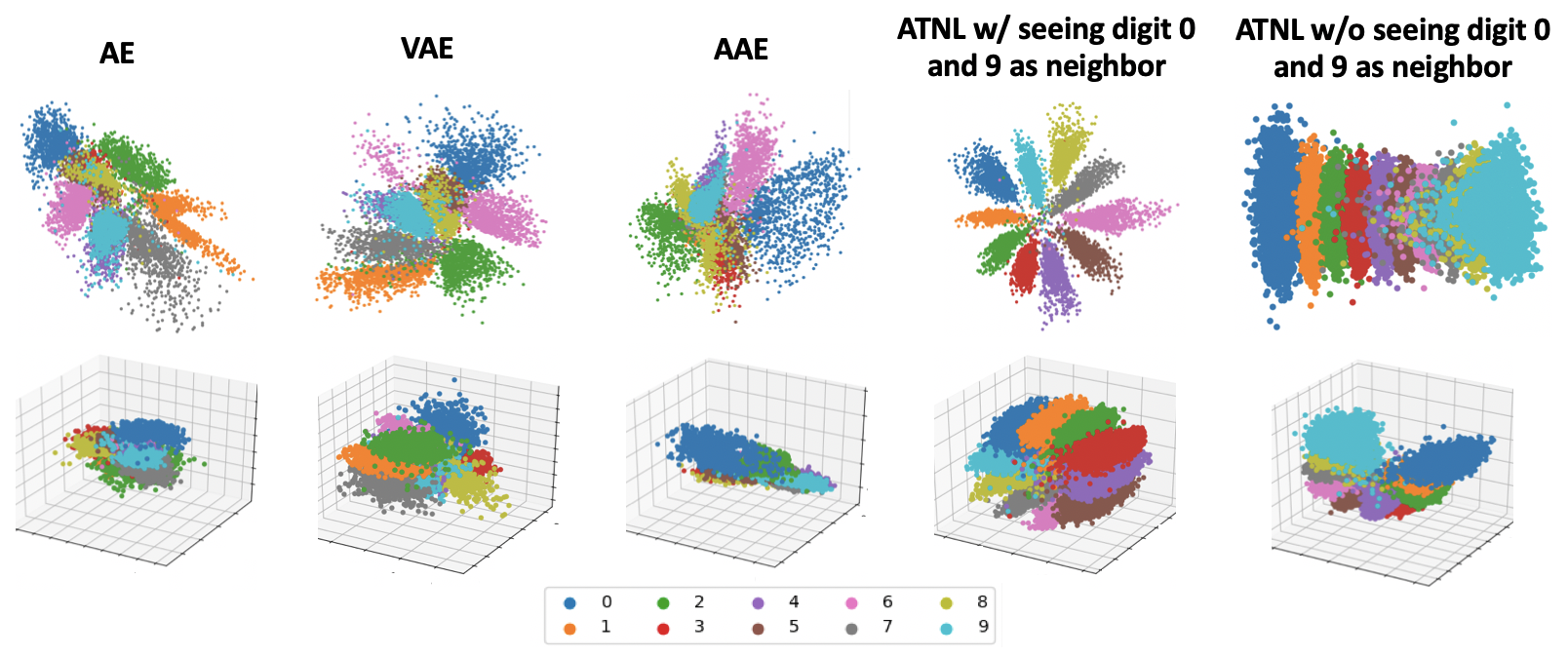}
  \caption{t-SNE visualization on MNIST. The first and second rows show 2D and 3D visualizations of $z$ produced by AE, VAE, AAE, ACAI and our ATNL, respectively.}
  \label{fig:mnist_tsne}
\end{figure}

\begin{figure}[t]
  \centering
  \includegraphics[width=\linewidth]{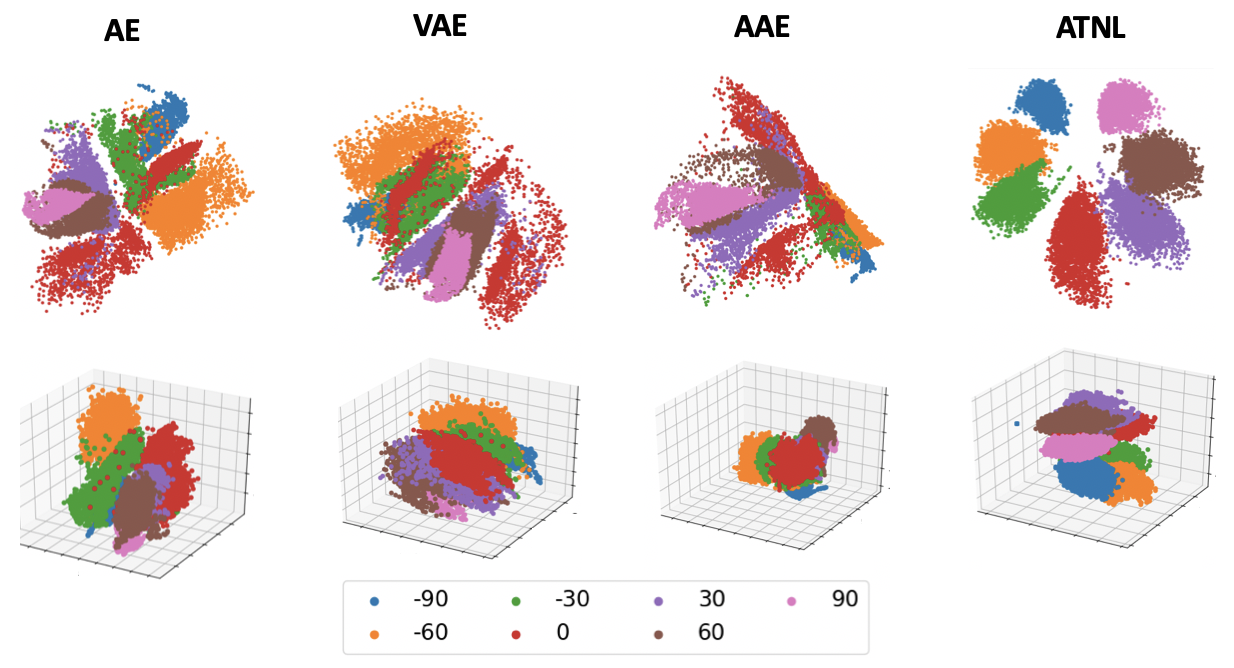}
%   \vspace{-3mm}
  \caption{t-SNE visualization on Multi-PIE. The first and second rows show 2D and 3D visualizations of $z$ produced by AE, VAE, AAE and our ATNL, respectively.}
  \label{fig:multipie_tsne}
%   \vspace{-3mm}
\end{figure}

\subsection{Visualization via t-SNE projection}

\paragraph{MNIST.}
We now visualize latent representation derived by AE, VAE, AAE, and VAE with our proposed ATNL on MNIST in Figure~\ref{fig:mnist_tsne}. We provide the results of both 2D t-SNE and 3D t-SNE results on the latent representation $z$ of the above models. It is important to note that, the representation $z$ produced by our ATNL (the last column) is properly clustered with respect to class of digits with task-oriented semantic information, showing that the derived latent representation by our ATNL has preserved the most perceivable semantic information. Once the ATNL loss is excluded (standard VAE), the resulted representation $z$ exhibits several overlapping parts shown in the second column.

\paragraph{CMU Multi-PIE.}
We further present qualitative results on the latent representations derived by AE, VAE, AAE and VAE with our proposed ATNL on CMU Multi-PIE in Figure~\ref{fig:multipie_tsne}. We provide the results of 2D t-SNE and 3D t-SNE results on the latent representation $z$ of the above models. Images belonging to same category were visualized in specific color. It is worth repeating that, the representation $z$ produced by our ATNL (the last column) on the CMU Multi-PIE is also properly clustered with respect to class of digits, demonstrating sufficient task-oriented semantic information.

\begin{figure}
\vspace{2mm}
  \centering
  \includegraphics[width=\linewidth]{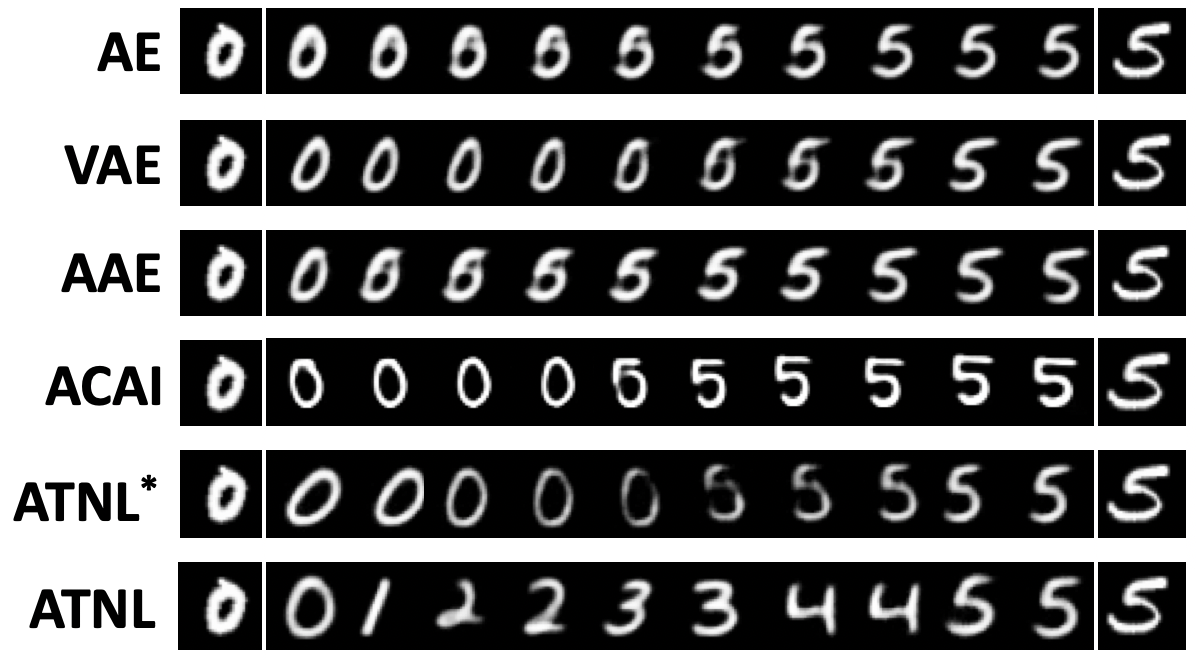}
  \caption{Image manipulation via linear or spherical semantic interpolation on MNIST. Given two input images in the first and last columns, we present the intermediate images of AE, VAE, AAE, ACAI, ATNL$^*$ and ATNL. Note that the first five models perform linear interpolation, while our ATNL performs the spherical one.
  }
  \label{fig:mnist_s0to5}
% \vspace{1mm}
\end{figure}

\subsection{Linear vs. Spherical Semantic Interpolation}

\begin{figure}
% \vspace{2mm}
  \centering
  \includegraphics[width=\linewidth]{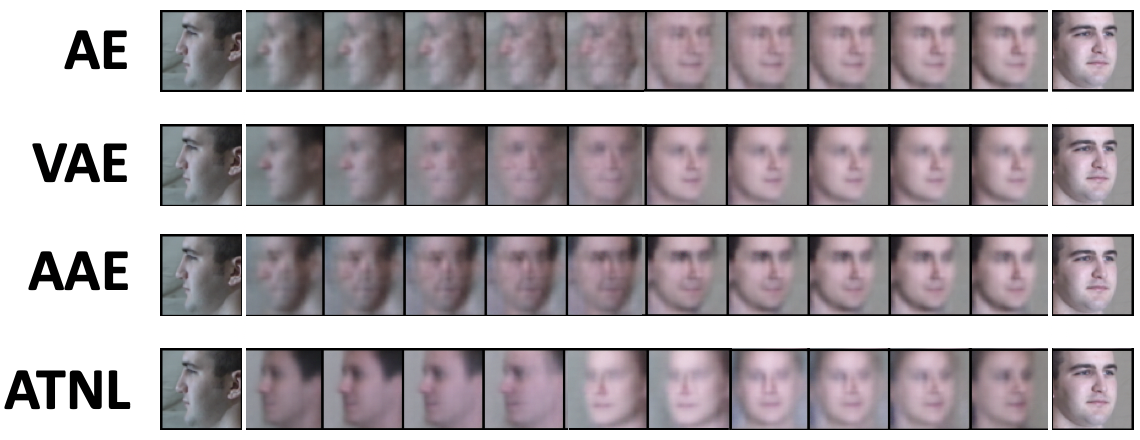}
  \caption{Image manipulation via linear or spherical semantic interpolation on Multi-PIE. Given two input images in the first and last columns, we present the intermediate images of AE, VAE, AAE, ACAI and our ATNL. Note that the first three models perform linear interpolation, while our ATNL perform the spherical one.
  }
  \label{fig:multipie_slerp}
  \vspace{1mm}
\end{figure}

We now further assess the ability of our ATNL in describing distribution of latent representation. Given two images of different digit as inputs, we compare image outputs produced by AE, VAE, AAE, ACAI, ATNL$^*$ with linear interpolation and our ATNL with spherical semantic interpolation in Figure~\ref{fig:mnist_s0to5}. As shown in the first to fifth rows in Figure~\ref{fig:mnist_s0to5}, AE, VAE, AAE, and ACAI, ATNL$^*$ simply performed linear interpolation of latent representations in manipulating output images, which were visually and gradually warped from digit 0 to digit 5. On the other hand, our ATNL was able to interpolate latent vectors producing images in semantics/digit order (i.e., digits $0 \rightarrow 1 \rightarrow 2 \rightarrow 3 \rightarrow 4$ interpolation for AE, VAE, AAE and spherical semantic interpolation for our ATNL. As shown in Figure~\ref{fig:multipie_slerp}, the output images were produced by interpolating face images in -90 degrees and those in 30 degrees. Similar to MNIST, our ATNL on Multi-PIE was also able to properly generate images in semantics/pose order, while previous models simply produced image outputs with visual appearance changes. From the above experiments, the ability of our ATNL in representation learning and describing task-oriented semantic information can be verified.

\begin{table}[t!]
\vspace{1mm}
\small
\centering
\caption{Classification performances on MNIST and CMU Multi-PIE using different models. Note that VAE-TL denotes the standard VAE with a supervised triplet loss.}
	\begin{tabular}{c||cc|cc}
		\toprule
		\multirow{2}{*}{Dataset} & \multicolumn{2}{c|}{Unsup.} & \multicolumn{2}{c}{Sup.}\\
		 & AE  & VAE & VAE-TL & \textbf{Ours} \\
		\midrule
		MNIST    & $94.63$  & $96.42$ & $98.46$ & \textbf{99.14} \\
		\midrule
		CMU Multi-PIE   & $84.80$  & $86.26$ & $91.36$ & \textbf{92.41} \\
		\bottomrule
	\end{tabular}
	%\vspace{10pt}
	\label{table_1}
%  	\vspace{1mm}
\end{table}

\subsection{Quantitative Evaluation}

We further conduct quantitative experiments for examining the effectiveness of our ATNL in representation learning. For fair comparison, we employ a simple one-layer classifier (1 linear layers followed by softmax) for two classification tasks. The number of class in MNIST is 10 for digit classification and the number of class in CMU Multi-PIE is 7 for pose classification. We adopt AE and VAE as our baseline models while we employ VAE with additional triplet loss (VAE-TL), as our comparative model. As shown in Table~\ref{table_1}, we report the classification performance on both MNIST and CMU Multi-PIE, which achieve $99.14\%$ and $92.41\%$ at accuracy, respectively. The experiment demonstrates that our designed Angular Triplet-Neighbor Loss (ATNL) yields the superiority in two classification tasks.

\begin{figure}[t]
%   \vspace{-3mm}
  \centering
  \includegraphics[width=\linewidth]{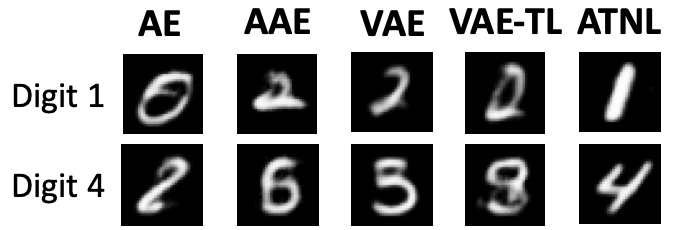}
  \caption{Examples of interpolated images on MNIST. For each digit image, it is produced by interpolating its two semantic neighbors (e.g., neighbors of digit 1 are digits 0 and 2, neighbors of digit 4 are digits 3 and 5). Note that AE, AAE, VAE, VAE-TL use linear interpolation while our ATNL applies spherical semantic interpolation.}
  \label{fig:m_inter_example}
  \vspace{1mm}
\end{figure}

\begin{figure}[t]
\vspace{2mm}
  \centering
  \includegraphics[width=0.7\linewidth]{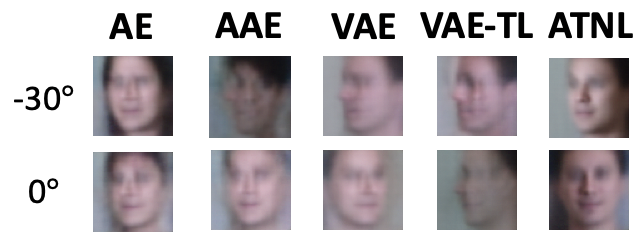}
  \caption{Examples of interpolated images on CMU Multi-PIE. For each pose image, it is produced by interpolating its two semantic neighbors (e.g., neighbors of -30 degrees are -60 degrees and 0 degrees, neighbors of 0 degrees are -30 degrees and 30 degrees). Note that AE, AAE, VAE, VAE-TL use linear interpolation while our ATNL applies spherical semantic interpolation.}
  \label{fig:cmu_inter_example}
  \vspace{-2mm}
\end{figure}

\subsection{Assessment of Interpolated Images}

\begin{figure}[t!]
  %\vspace{-3mm}
  \centering
  \includegraphics[width=\linewidth]{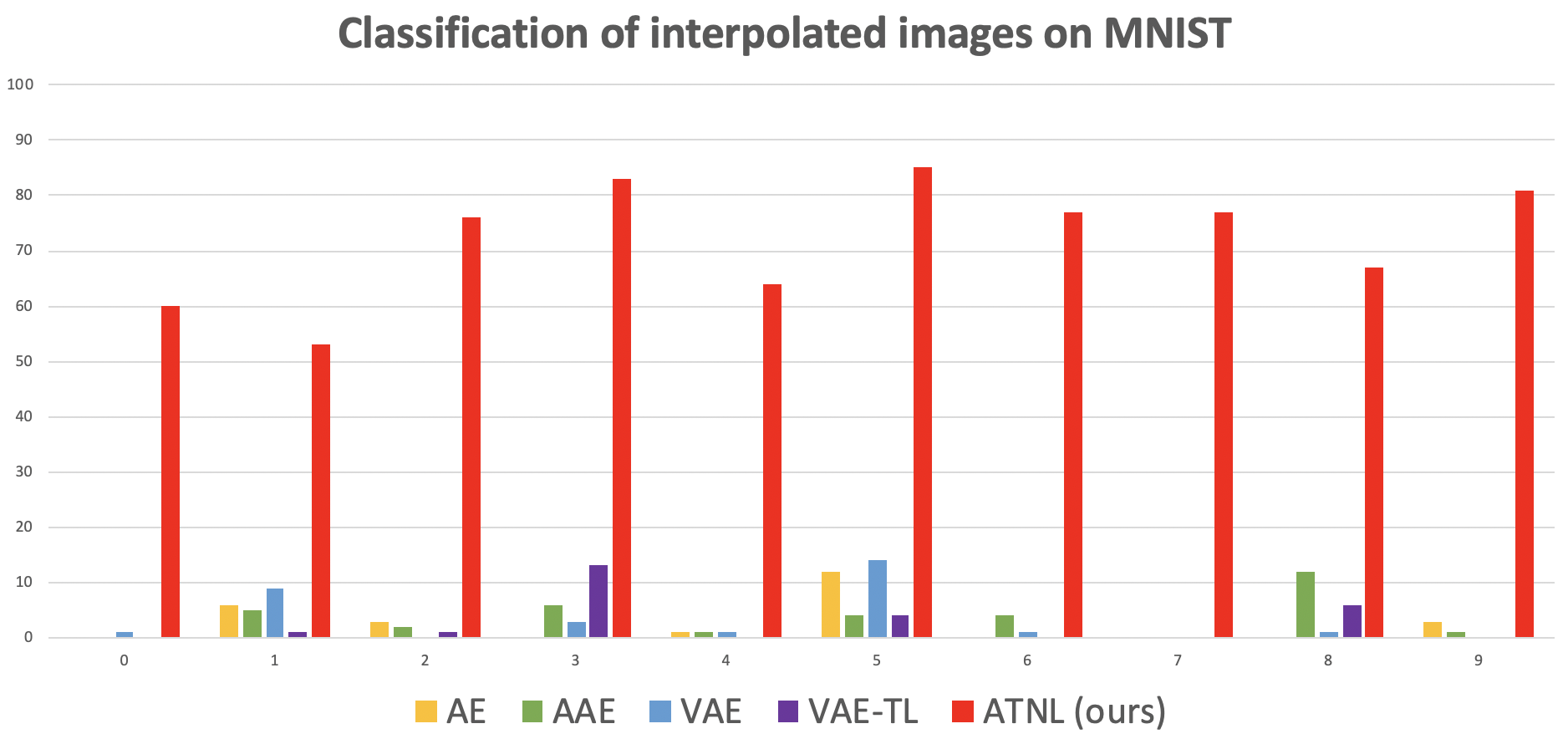}
  \caption{Classification of interpolated images by linear/spherical semantic interpolation on MNIST (i.e., Figure~\ref{fig:m_inter_example}). Note that only images produced by our spherical semantic interpolation are able to be recognized with satisfactory performances.}
  \label{fig:m_inter_class}
  \vspace{-2mm}
\end{figure}

\begin{figure}[t!]
  \centering
  \includegraphics[width=\linewidth]{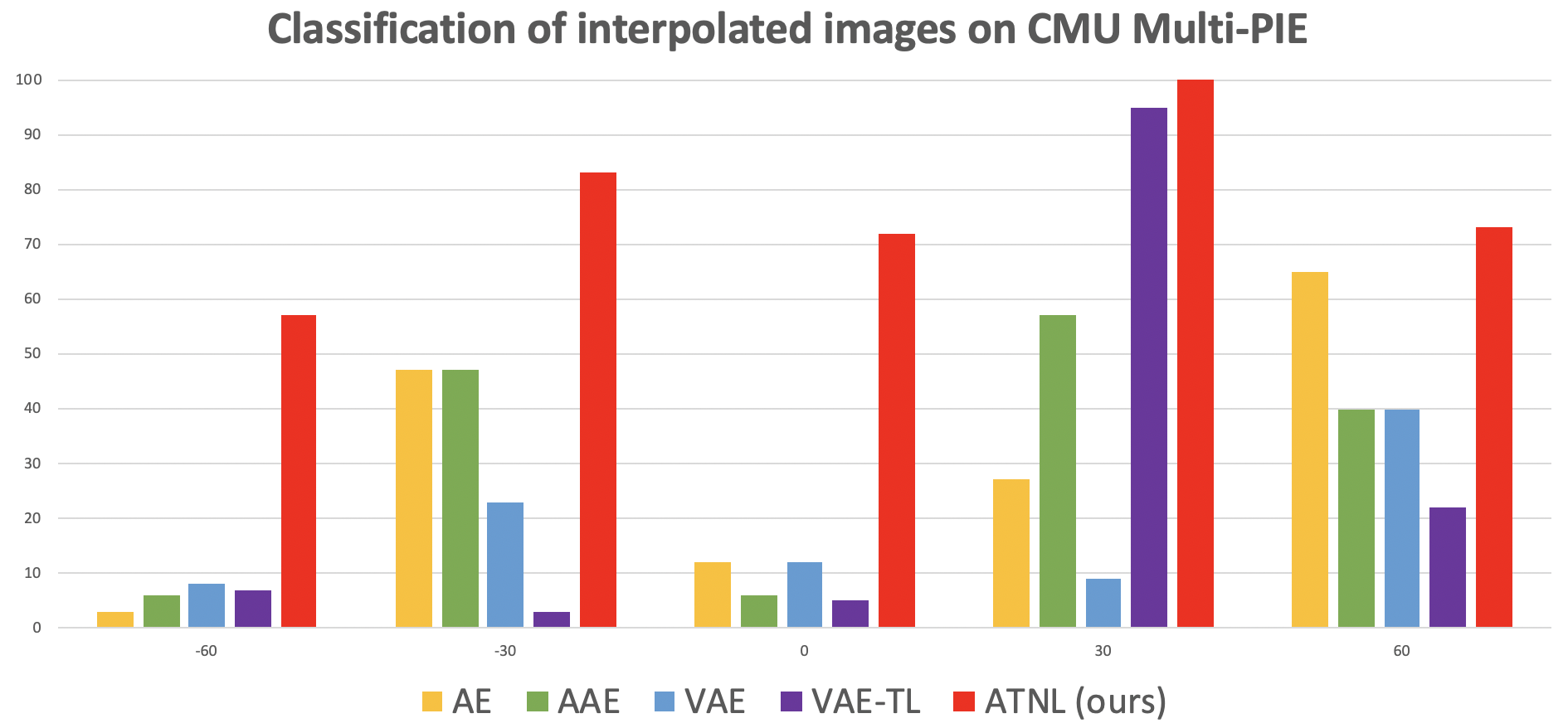}
  \caption{Classification of interpolated images by linear/spherical semantic interpolation on Multi-PIE (i.e.,~\ref{fig:cmu_inter_example}). Note that only images produced by our spherical semantic interpolation are able to be recognized with satisfactory performances.}
  \label{fig:cmu_inter_class}
%   \vspace{-1mm}
\end{figure}

We now qualitatively and quantitatively evaluate the interpolated images synthesized by our ATNL. We adopt linear interpolation for AE, AAE, VAE and VAE-TL and spherical semantic interpolation for our ATNL, and assess the synthesized images are interpolated by their semantic neighbors (e.g., the neighbors of digit 1 are digits 0 and 2 for MNIST). The example interpolated images are shown in Figure~\ref{fig:m_inter_example} and Figure~\ref{fig:cmu_inter_example}, respectively.
Also, as shown in Figure~\ref{fig:m_inter_class} and Figure~\ref{fig:cmu_inter_class}, we report the classification performance on such interpolated images. From the above figures, we see that the images synthesized by our ATNL were visually practical and allowed very promising recognition performances.

\subsection{Extension as Data Hallucination for Few-Shot Learning}
We further apply our learned representation with spherical semantic interpolation as a data hallucination technique for few-shot learning (FSL). That is, instead of applying standard data augmentation techniques, we interpolate images from their semantic neighbors to increase data for categories with few training samples. Both K-nearest-neighbor (KNN) and standard Convolution Neural Networks (CNN) are applied as classifiers. As shown in Table~\ref{table_2}, we achieved the highest ac curacies on MNIST, which were clearly superior to those reported by baseline and standard data augmentation techniques. This confirms that, when the semantic labels are known, our proposed model can be properly served as a data hallucination technique for FSL tasks.

\begin{table}[t!]
\small
\vspace{1mm}
\centering
\caption{Few-shot classification on MNIST with different data hallucination techniques. Note that the baseline denotes uses of image scale/brightness/cropping variants for data hallucination, while ours applies our semantically interpolated images as hallucinated data (e.g.,~\ref{fig:m_inter_example}).}
	\begin{tabular}{c||c|c}
		\toprule
		\multirow{2}{*}{Method}&
		\multicolumn{2}{c}{Classifiers} \\
		& KNN & CNN\\
		\midrule
		w/o data hallucination & $61.5$& $86.3$\\
		\midrule
		w/ data augmentation (baseline) & $62.4$ & $86.8$\\
		\midrule
		w/ ATNL data augmentation (ours) & \textbf{78.7}& \textbf{94.4}\\
		\bottomrule
	\end{tabular}
	\label{table_2}
 	\vspace{-1mm}
\end{table}

\begin{figure}[t!]
%\vspace{-3mm}
  \centering
  \includegraphics[width=\linewidth]{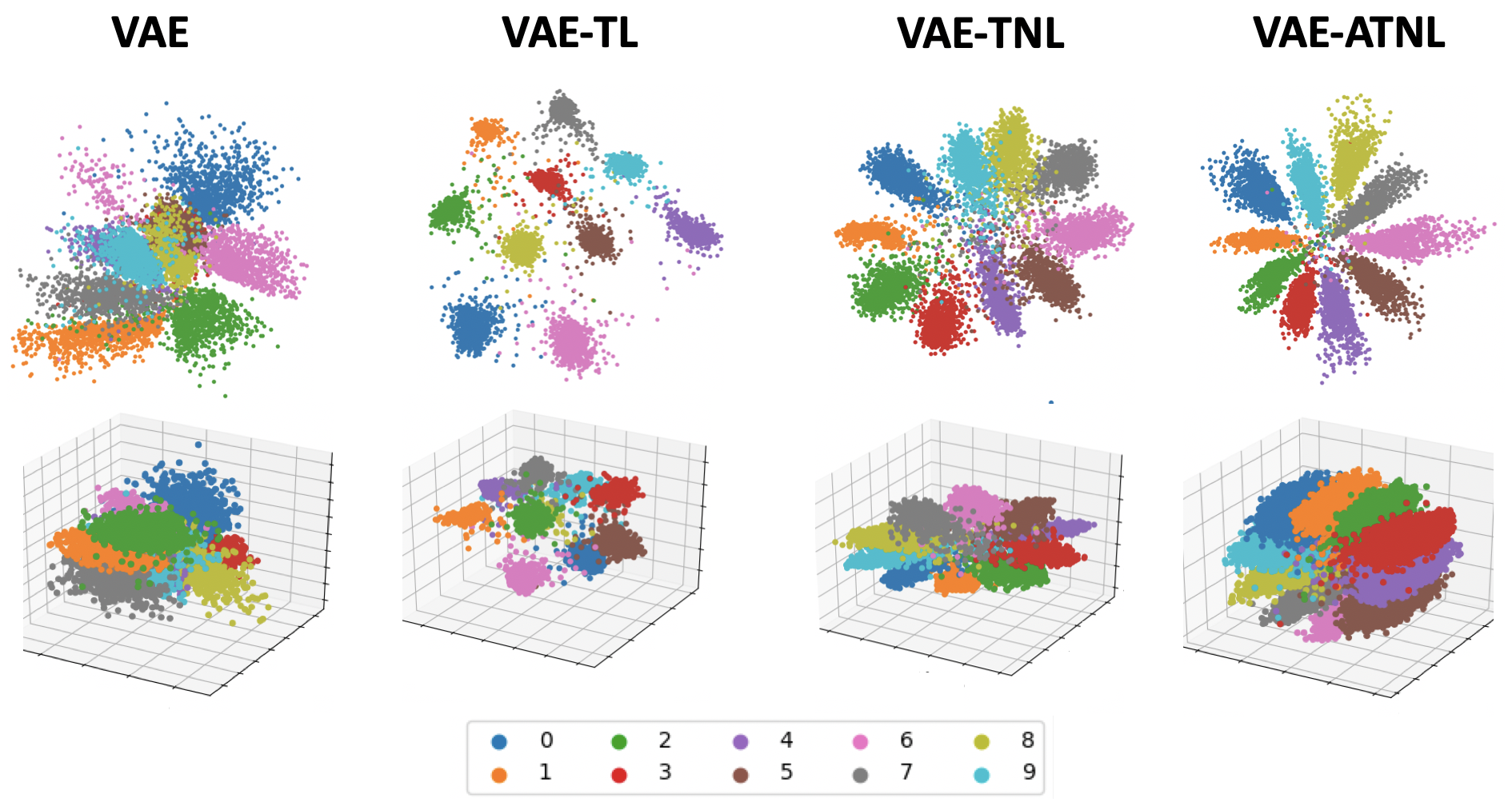}
  \caption{2D/3D t-SNE visualization on variants of our model using MNIST. We visualize  $z$ produced by VAE, VAE-Triplet Loss (VAE-TL), VAE-Triplet Neighbor Loss (VAE-TNL) and our VAE-ATNL. Note that each color represents a specific digit class.}
  \label{fig:ablation_tsne}
%   \vspace{-3mm}
\end{figure}

\subsection{Ablation Studies}

We visualize latent representations derived by VAE, VAE with triplet loss (VAE-TL), VAE with triplet loss based on our semantically positive/negative neighbors (VAE-TNL), and VAE-ATNL (ours) using MNIST in Figure~\ref{fig:ablation_tsne}. By both 2D and 3D t-SNE results, we see that $z$ of the same digit derived from VAE-TNL and VAE-ATNL were properly clustered while the clustering results exhibited task-oriented semantic information. Finally, $z$ derived by VAE-ATNL (ours) were mapped on to a spherical hyperplan with better separation between classes, showing that the derived latent representation by VAE-ATNL (ours) has preserved the most perceivable semantic information and geometric interpretation.
%\vspace{2em}
\section{Conclusion}

In this paper, we addressed the learning of interpretatble and interpolatable latent representations of visual data. By observing task-oriented semantics information, we proposed Angular Triplet-Neighbor Loss (ATNL) can be applied to existing generative models like VAE and have the latent representation distributions fit the associated semantic information. Together with the presented spherical semantic interpolation, we are able to synthesize images with particular categories of interest. In the experiments, we provided qualitative and quantitative results on two benchmark datasets, along with ablation studies and visualization to verify the effectiveness of our work. Moreover, we further apply our model as a data hallucination technique for few-shot learning, which is a practical yet challenging tasks in computer vision.

\section*{Acknowledgments}
%\noindent\textbf{Acknowledgments}\\
This work was supported in part by the Ministry of Science and Technology of Taiwan under grant MOST 109-2634-F-002-037.
\bibliographystyle{ieee}
\bibliography{egbib}

\end{document}